\documentclass[runningheads]{llncs}

\usepackage{eccv}

\usepackage{eccvabbrv}

\usepackage{graphicx}
\usepackage{booktabs}
\usepackage[accsupp]{axessibility}
\usepackage[table,xcdraw]{xcolor}
\usepackage[breaklinks,colorlinks,urlcolor=eccvblue,citecolor=eccvblue]{hyperref}
\usepackage{tikz}
\usetikzlibrary{positioning,arrows.meta,fit,calc}
\usepackage[utf8]{inputenc}
\usepackage[T1]{fontenc}
\usepackage{url}
\usepackage{booktabs}
\usepackage{amsfonts}
\usepackage{nicefrac}
\usepackage{microtype}
\usepackage{amsmath} 
\usepackage{graphicx}
\usepackage{array}
\usepackage{booktabs}
\usepackage[normalem]{ulem}
\useunder{\uline}{\ul}{}
\usepackage{comment}
\usepackage{wasysym}
\usepackage{multirow}
\usepackage{subcaption}
\usepackage{amssymb}
\usepackage{makecell}
\usepackage{pifont}

\usepackage{cuted}
\usepackage{multirow}
\usepackage{textcomp}
\usepackage[group-separator={,}, output-decimal-marker={.}, group-minimum-digits=4, group-digits=integer]{siunitx}

\usepackage{orcidlink}

\usepackage{adjustbox}
\newcommand{\cmark}{\ding{51}}
\newcommand{\xmark}{\ding{55}}

\newcolumntype{g}{>{\columncolor{lightgray}}c}
\definecolor{lightgreen}{rgb}{0.88, 1, 0.88}
\newcolumntype{g}{>{\columncolor{lightgreen}}c}
\newcolumntype{g}{>{\columncolor{lightgray}}c}
\definecolor{rowcolor}{rgb}{0.918, 0.918, 0.918}
\newcolumntype{g}{>{\columncolor{lightblue}}c}
\definecolor{lightorange}{rgb}{1.0, 0.94, 0.80}
\newcolumntype{g}{>{\columncolor{lightorange}}c}

\newcommand{\method}{SafeLand}
\newcommand{\rotheadL}[1]{\rotatebox[origin=l]{90}{#1}}

\DeclareMathOperator*{\argmax}{arg\,max}

\begin{document}

\title{\method: Safe Autonomous Landing in Unknown Environments with Bayesian Semantic Mapping} 

\titlerunning{\method}

\author{Markus Gross\textsuperscript{1,2,3,$\star$},\quad Andreas Greiner\textsuperscript{1},\quad Sai B. Matha\textsuperscript{1},\\Felix Soest\textsuperscript{1},\quad Daniel Cremers\textsuperscript{2,3},\quad Henri Meeß\textsuperscript{1}
}

\authorrunning{M.~Gross et al.}

\institute{\textsuperscript{1}Fraunhofer Institute IVI \quad \textsuperscript{2}TU Munich \quad \textsuperscript{3}Munich Center for Machine Learning}

\maketitle

\begingroup
\renewcommand{\thefootnote}{}
\footnotetext{$^\star$Corresponding author: markus.gross@tum.de.}
\endgroup

\begin{abstract}
    Autonomous landing of uncrewed aerial vehicles (UAVs) in unknown, dynamic environments poses significant safety challenges, particularly near people and infrastructure, as UAVs transition to routine urban and rural operations.
    Existing methods often rely on prior maps, heavy sensors like LiDAR, static markers, or fail to handle non-cooperative dynamic obstacles like humans, limiting generalization and real-time performance.
    To address these challenges, we introduce \method, a lean, vision-based system for safe autonomous landing (SAL) that requires no prior information and operates only with a camera and a lightweight height sensor.
    Our approach constructs an online semantic ground map via deep learning-based semantic segmentation, optimized for embedded deployment and trained on a consolidation of seven curated public aerial datasets (achieving 70.22\% mIoU across 20 categories), which is further refined through 
    Bayesian probabilistic filtering with temporal semantic decay to robustly identify metric-scale landing spots. 
    A behavior tree then governs adaptive landing, iteratively validates the spot, and reacts in real time to dynamic obstacles by pausing, climbing, or rerouting to alternative spots, maximizing human safety.
    We extensively evaluate our method in 200 simulations and 60 end-to-end field tests across industrial, urban, and rural environments at altitudes up to 100m, demonstrating zero false negatives for human detection.
    Compared to the state of the art, SafeLand achieves sub-second response latency, substantially lower than previous methods, while maintaining a superior success rate of 95\%.
    To facilitate further research in aerial robotics, we release \method's segmentation model as a plug-and-play ROS package, available at \url{https://github.com/markus-42/SafeLand}.
\end{abstract}

\section{Introduction}
\label{sec_intro}

Autonomous landing of uncrewed aerial vehicles (UAVs) is a safety-critical task, especially in proximity to people and infrastructure~\cite{Tovanche_Picon2024,farajijalal2025safety}. As UAVs move from isolated use cases toward dense deployments in logistics, inspection, and emergency response~\cite{sssurvey}, regulatory frameworks such as the Specific Operations Risk Assessment (SORA)~\cite{farajijalal2025safety} impose stringent risk constraints. Landing systems must therefore ensure safety under diverse and only partially known conditions while remaining lightweight and cost-efficient. Software and vision-based methods are central to this development, as they offer precise relative positioning from compact, low-power sensors, and are applicable across a multitude of UAV platforms~\cite{sssurvey,farajijalal2025safety}.\\

To this end, most vision-centric methods for \textbf{\emph{safe autonomous landing (SAL)}} are marker-based~\cite{sssurvey}: they achieve high accuracy and robustness on prepared landing pads by exploiting carefully designed markers and controlled infrastructure~\cite{11425768}.
However, these systems assume instrumented, static landing sites and thus do not generalize to generic outdoor operations, such as emergency landings in populated areas, or slowly changing environments, such as agricultural fields.
Moreover, beyond marker-based methods, existing works employ markerless, hybrid, and multi-sensor approaches, which exhibit systematic limitations~\cite{farajijalal2025safety}: many methods (i) assume prior maps, (ii) lack explicit metric reasoning due to the scale-ambiguous nature of camera-based projective geometry, (iii) depend on heavy and power-consuming depth sensors such as LiDAR or radar, (iv) treat humans as generic obstacles if at all, and (v) are rarely validated end-to-end in sophisticated field experiments.\\

\begin{figure}[!t]
    \centering
    \includegraphics[width=.8\linewidth]{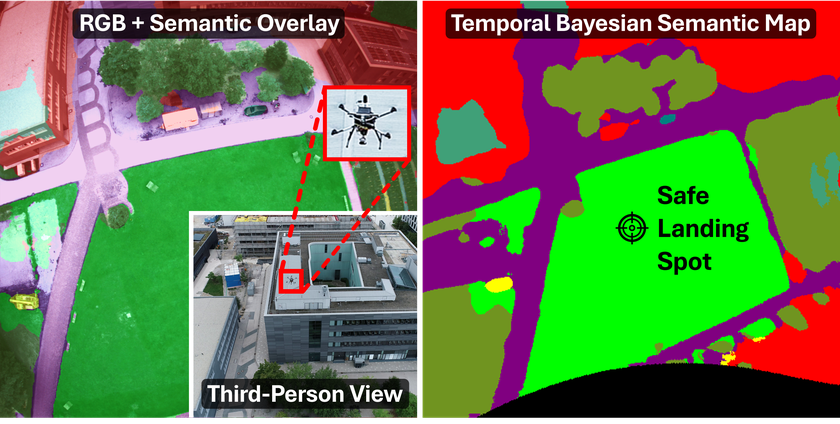}
    \caption{\method~predicts a semantic map, projects it to a metric ground plane, filters it probabilistically over time to select a landing spot, and uses a behavior tree for safe real-time landing in unknown and dynamic environments.}
    \label{fig_thumbnail}
\end{figure}

Motivated by these gaps, we introduce \emph{\textbf{\method}}~(Fig.~\ref{fig_thumbnail}), a lightweight, vision-based SAL system that builds an online semantic ground map from deep learning-based semantic segmentation and height-above-ground-level sensing, refines it via Bayesian temporal–probabilistic filtering, and uses a behavior tree to adapt landing decisions to non-cooperative dynamic obstacles, including people.
Since generic vision foundation models underperform on aerial semantic segmentation~\cite{sam_remote_sensing}, we train a network on curated public aerial datasets and release it as a plug-and-play ROS package for the community.

\vspace{0.5em}
\noindent In summary, our contributions are as follows:

\begin{itemize}
    \item We present \method, a lightweight, vision-based UAV landing solution that operates with a minimal sensor setup, requires no prior information about the landing site, and ensures safety for humans and infrastructure.
    \item We train a robust aerial semantic segmentation model from $7$ curated datasets, achieving \SI{70.22}{mIoU}, which we further optimize for embedded deployment using the TensorRT engine.
    \item We evaluate \method~in 200 simulations and 60 diverse field tests across industrial, urban, and rural environments at \SI{25}{m} to \SI{100}{m} altitudes, with deliberate interventions by non-cooperative dynamic obstacles, such as humans.
    \item \method~substantially outperforms baselines with sub-second response latency and a \SI{95}{\%} landing success rate.
    \item Upon publication, we will release \method's segmentation model as a plug-and-play ROS package to facilitate further research in vision-based aerial robotics.
\end{itemize}

\section{Related Work}
\label{sec_related_work}

Apart from methods that rely on depth sensors, such as LiDAR and radar~\cite{10971405,hub_1,hub_1_2,neves2024,lidar_1,lidar_2,monodepthEML}, which often add significant weight and power consumption, vision-based methods have gained prominence as lightweight, cost-effective alternatives for SAL in unknown environments.
These vision-centric approaches process camera imagery to detect suitable landing sites, and are often categorized as traditional image processing for feature extraction, such as edge detection or texture analysis~\cite{canny_land,free_1,freelsd}, supervised zone-classification based on visual cues~\cite{satellite,gabor,SORA,safeeye_3,pakistani}, or more advanced deep learning-driven semantic segmentation to delineate terrain types~\cite{landing_target,11247278,russian_paper,safeeye_1,safeeye_2}.
A prevalent limitation across many is the dependence on priors, such as pre-existing maps of potential landing areas~\cite{hub_2,multimodal,PIECZYNSKI2024107864,elpodense} or structured environments with instrumented landing pads~\cite{hub_1,hub_1_2,neves2024}, which restricts their applicability to truly dynamic or unprepared scenarios where environmental changes or lack of infrastructure demand greater flexibility~\cite{farajijalal2025safety}.
Furthermore, to address the inherent scale ambiguity in camera-based projective geometry, several methods integrate metric compensation strategies, including assumptions of flat ground for homography-based transformations~\cite{safe2ditch,luczak2025autonomous}, depth cameras~\cite{luczak2025autonomous}, or fusion with complementary data sources like inertial measurements and altitude sensors to recover absolute scales~\cite{landing_target,russian_paper,pakistani,canny_land,free_1,freelsd,circles_2,circles_3,ESLS}, enabling precise sizing of landing spots relative to UAV dimensions.\\

\indent Importantly, safety considerations, particularly for people as non-cooperative dynamic obstacles, are underexplored in the literature and, when addressed, are typically handled only in a subset of works through techniques like population density estimation from aerial views~\cite{circles_1,elpodense} or real-time visual tracking to monitor and predict human movements~\cite{circles_2,De_La_Torre_Vanegas_2026,circles_3,ESLS,PIECZYNSKI2024107864,safe2ditch}.
Notably, comprehensive validation remains a gap~\cite{farajijalal2025safety}: to the best of our knowledge, only Safe2Ditch~\cite{safe2ditch} conducts sophisticated end-to-end field tests under real-world conditions, demonstrating practical functionality with human interference, whereas the majority evaluate individual components, such as standalone perception modules, or confine assessments entirely to simulations~\cite{safeeye_1,safeeye_2,safeeye_3,monodepthEML,satellite,gabor,SORA,lidar_1,lidar_2,pakistani,canny_land,free_1,freelsd,PIECZYNSKI2024107864,elpodense,circles_1,circles_2,circles_3,ESLS,hub_1,hub_1_2,neves2024,hub_2,multimodal,landing_target,russian_paper,farajijalal2025safety}, limiting insights to isolated challenges, such as lighting conditions,
without considering holistic system viability.\\

\indent In summary, \method~follows this line of vision-based SAL but differs in three key aspects. It trains a dedicated aerial segmentation model from complementary datasets to handle diverse unknown environments, restores metric scale using only a lightweight height above ground level (AGL) sensor instead of heavy depth hardware or prior maps, and explicitly targets human safety with conservative Bayesian filtering of person detections, validated in 200 simulations and 60 field tests across industrial, urban, and rural sites up to \SI{100}{m} AGL.

\section{Methodology}\label{sec_methodology}

\begin{figure*}[t!]
	\centering
	\includegraphics[width=.98\textwidth]{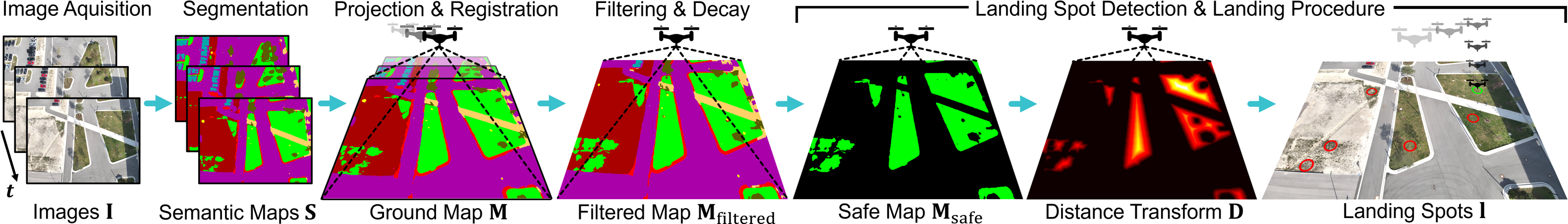}
	\caption{Chronological processing steps of \method. An overview is provided in Sec.~\ref{subsec_method_overview}.}
	\label{fig_pipeline}    
\end{figure*}

\subsection{Method Overview}\label{subsec_method_overview}

\method~(see Fig.~\ref{fig_pipeline}) uses only a nadir-pointing camera and an AGL sensor. It applies deep learning-based semantic segmentation on temporal camera frames, projects the resulting semantic probabilities onto a discrete, metrically scaled ground plane to obtain a semantic ground map, and iteratively filters this map via Bayesian inference and semantic decay for robust perception. We then extract safe landing spot candidates, and a behavior tree-based landing procedure navigates the UAV toward the selected spot while re-selecting a new spot if the current one becomes invalid. This pipeline continuously evaluates the environment, reacts in real time to non-cooperative obstacles, and operates without prior map information or heavy sensors, supporting generic deployment.

\subsection{Image Acquisition and Semantic Segmentation}\label{subsec_segmentation}

A nadir-pointing RGB camera captures real-time imagery for semantic segmentation.
Robust perception in previously unseen and diverse terrain requires diverse training data, yet existing aerial datasets are individually too narrow, and current vision foundation models for semantic segmentation underperform in the aerial domain~\cite{sam_remote_sensing}.
We therefore consolidate seven open-source datasets~\cite{aeroscapes,floodnet,icg,swiss-okutama,UAVid,udd6,vdd} that provide aerial imagery with semantic segmentation ground-truth and curate their labels into 20 unified classes.

We train a SegFormer MiT-B3 model~\cite{segformer} with \SI{47.3} million parameters and obtain \SI{70.22}{\%} mIoU, detailed in Tab.~\ref{tab_semantic_segmentation}.
We select SegFormer for its balance between accuracy and efficiency, which enables deployment on resource-constrained onboard hardware.
This model is further optimized for embedded deployment using the TensorRT inference engine~\cite{tensorrt} with FP16 precision calibration, increasing standalone throughput to \SI{12}{FPS}.

Formally, let $\mathcal{C}$ be the set of semantic classes summarized in Tab.~\ref{tab_semantic_segmentation}. The camera image $\mathbf{I} \in \mathbb{R}^{M \times N \times 3}$ is segmented as
\begin{equation}
    \mathbf{S} = \Phi_s(\mathbf{I}) \; ,
\end{equation}
where $\Phi_s$ denotes the segmentation model, $\mathbf{S} \in \mathbb{R}^{M \times N \times C}$ is the predicted map, $M \times N$ represents the image resolution, and $C = \lvert\mathcal{C} \lvert$ is the number of semantic classes.

Overall, semantic segmentation relies only on a lightweight and cost-efficient camera, removes the need for prior map information, facilitates generalization to diverse environments, and improves safety by explicitly modeling humans.

\subsection{Projection and Registration}\label{subsec_projection_registration}

These steps follow the method described in \cite{eml_paper}, and assume a flat ground and AGL measurements.
In summary, the undistorted semantic map is projected onto a ground plane using the UAV's pose and height, in addition to camera intrinsics and extrinsics, yielding a discrete semantic ground map $\mathbf{M} \in \mathbb{R}^{X \times Y \times C}$ with fixed size and metric scale.

Formally, a homogeneous pixel coordinate $\mathbf{s} \in \mathbb{P}^2$ on the predicted semantic map $\mathbf{S}$ and its corresponding homogeneous 3D point $\tilde{\mathbf{m}} \in \mathbb{P}^3$ on the semantic ground map $\mathbf{M}$ are related by
\begin{equation}
    \label{eq_cam_to_ground}
    \mathbf{s} \;=\;
    \left[ K \mid \mathbf{0} \right]
    \cdot
    \mathbf{T}_{\text{cam}\leftarrow\text{world}}
    \cdot
    \tilde{\mathbf{m}}\; ,
\end{equation}
where $\mathbf{0} \in \mathbb{R}^3$ is the null vector, $\mathbf{K} \in \mathbb{R}^{3\times3}$ are camera intrinsics, and $\mathbf{T}_{\text{cam}\leftarrow\text{world}} \in \text{SE(3)}$ denotes a rigid transformation from planar world coordinates to camera coordinates \cite{eml_paper}.

Subsequent registration aligns earlier maps with the current one, by transforming a Cartesian point $\mathbf{m}_{t-1} \in \mathbb{R}^3$ via
\begin{equation}
\label{eq:map_registration}
\mathbf{M}_{t}\!\bigl(\mathbf{m}_{t-1}\bigr)
\;=\;
\mathbf{M}_{t-1}\!\bigl(\mathbf{H}_{\,t\leftarrow (t-1)}\cdot\mathbf{m}_{t-1}\bigr)\;,
\end{equation}
where $\mathbf{M}_{t}$ and $\mathbf{M}_{t-1}$ denote the ground maps at times $t$ and $(t-1)$, respectively. $\mathbf{H}_{\,t\leftarrow (t-1)} \in \mathbb{R}^{3\times3}$ is a homography that warps points from the previous map’s coordinate system to the current one, which is computed from the camera extrinsics at times $t$ and $(t-1)$.
In practice, four or more corresponding points are sufficient to estimate this homography whenever the extrinsics change.
For more details, see~\cite{eml_paper}.

\subsection{Bayesian Filtering and Semantic Decay}\label{subsec_filtering_decay}

The segmentation model's predictions remain subject to residual errors and dataset-induced biases.
We therefore iteratively refine the semantic ground map $\mathbf{M}$ over time, enhancing robustness and safety of subsequent landing spot detection.

Formally, we apply Bayesian inference and treat the model's softmax outputs as unnormalized conditional class probabilities that serve as proxies for likelihood terms in the Bayesian update.
This filtering process updates the belief about the state of each pixel in the semantic ground map based on new observations from the segmentation model $\Phi_s$. 
Let $P(c \mid \mathbf{M}_t)$ be the posterior probability of class $c$ given the segmentation map $\mathbf{M}_t$ at time $t$.
The filtering process is expressed as
\begin{equation}
    P(c \mid \mathbf{M}_t) = \frac{P(\mathbf{M}_t \mid c) \cdot P(c \mid \mathbf{M}_{t-1})}{P(\mathbf{M}_t)} \; ,
\end{equation}
where $P(\mathbf{M}_t \mid c)$ represents the likelihood of observing the semantic ground map $\mathbf{M}_t$ given class $c$, $P(c \mid \mathbf{M}_{t-1})$ is the prior probability of class $c$ at time $t-1$, and $P(\mathbf{M}_t)$ is the evidence (marginal likelihood). While the evidence is often omitted when only relative posterior scores are needed, we compute it explicitly by normalizing the posterior to ensure numerical stability. We initialize with a uniform prior.\\
\indent Subsequently, semantic decay incrementally updates the posterior beliefs over time, incorporating new posteriors by applying a decay factor to downweight the contribution of past observations. Let $\alpha$ be the decay factor, and $\mathbf{M}_t$ be the refined semantic ground map at time $t$. This update is formulated as
\begin{equation}
    \label{eq_decay}
    \mathbf{M}_t(\mathbf{m},c) = \alpha \cdot \mathbf{M}_{t-1}(\mathbf{m},c) + (1 - \alpha) \cdot P\bigl(c \mid \mathbf{M}_t(\mathbf{m},c)\bigr) \; ,
\end{equation}
where $\mathbf{M}_t(\mathbf{m},c)$ and $\mathbf{M}_{t-1}(\mathbf{m},c)$ are the probabilities of class $c$ at position $\mathbf{m}$ in the refined map at time $t$ and $(t-1)$, respectively, and $\alpha$ is the decay factor, which is empirically set to $\alpha=0.1$. Eqn. \ref{eq_decay} also implies that, if no updates are received for $\alpha^{-1}$ iterations, the semantic estimate decays to $0$, ensuring that stale evidence is forgotten and the map remains up to date with recent observations.\\
\indent Furthermore, to ensure safety in our proposed method, we treat human detections conservatively by setting the class probability of any predicted person to $P(\mathbf{M}_t \mid c)=1$, where $c \;\hat{=}\;\text{\textit{person}}$. Moreover, semantic decay for persons is only applied if a different semantic class is predicted at the same position, ensuring that their probability does not decay to $0$ when no updates are received.\\
\indent Formally, the filtered map $\mathbf{M}_{\text{filtered}} \in \mathbb{R}^{X\times Y \times 1}$ is determined by selecting the class with the highest probability at each location $\mathbf{m}$ as the definitive class for that location:
\begin{equation}
    \label{eq_M_filtered}
    \mathbf{M}_{\text{filtered}}(\mathbf{m}) = \arg\max_{c} \mathbf{M}_t(\mathbf{m}, c) \; .
\end{equation}
By applying Bayesian filtering with semantic decay, we mitigate segmentation errors, ensure robust performance in unknown environments, and enhance human safety.

\begin{table}[t]
\centering
\caption{Intersection over union (IoU) and mean IoU (mIoU) in [\%] for semantic segmentation (Sec.~\ref{subsec_segmentation}). To ensure fair comparison, we retain the train/val/test splits of the original datasets~\cite{aeroscapes,floodnet,icg,swiss-okutama,UAVid,udd6,vdd}. Thus, some classes appear only in the training split and are absent from val/test, so their performance cannot be quantified and is marked as n/a. However, for these classes, we (1) qualitatively assess them and (2) confirm their high practical performance by successfully landing on them in real-world field tests (Sec.~\ref{subsec_field_experiments}). The lower Bicycle score is consistent with its sparse occurrence in the data. }
\renewcommand{\arraystretch}{0.8}
\resizebox{\linewidth}{!}{
\begin{tabular}{c|*{20}{c|}}
\toprule
{\textbf{mIoU}} &
\rotheadL{Road} &
\rotheadL{Dirt} &
\rotheadL{Gravel} &
\rotheadL{Rock} &
\rotheadL{Grass} &
\rotheadL{Vegetation} &
\rotheadL{Tree} &
\rotheadL{Obstacle} &
\rotheadL{Animal} &
\rotheadL{Person} &
\rotheadL{Bicycle} &
\rotheadL{Vehicle} &
\rotheadL{Water} &
\rotheadL{Boat} &
\rotheadL{Wall} &
\rotheadL{Roof} &
\rotheadL{Sky} &
\rotheadL{Drone} &
\rotheadL{Train-Track} &
\rotheadL{Background} \\
\midrule
{$\mathbf{70.22}$} &
{$92.85$} &
{n/a} &
{n/a} &
{n/a} &
{$89.18$} &
{$96.41$} &
{$81.52$} &
{$39.14$} &
{$81.74$} &
{$70.09$} &
{$1.88$} &
{$69.18$} &
{$83.30$} &
{n/a} &
{$83.49$} &
{$81.05$} &
{$97.40$} &
{$35.49$} &
{$50.61$} &
{$-$} \\
\bottomrule
\end{tabular}
}
\label{tab_semantic_segmentation}
\end{table}

\subsection{Landing Spot Detection}\label{subsec_landing_spot_detection}

A safe landing spot is identified in three steps (see Fig.~\ref{fig_pipeline}).
First, we define and rank semantic classes suitable for safe landing (e.g., grass, dirt, gravel), denoted as $\mathcal{C}_{\text{safe}}$, and only consider these classes in the safe semantic ground map $\mathbf{M}_{\text{safe}}$:
\begin{equation}
    \mathbf{M}_{\text{safe}} = \bigl\{\mathbf{M}_{\text{filtered}}(c) \mid c \in \mathcal{C}_{\text{safe}}\bigr\} \; .
\end{equation}
We then apply a distance transform \cite{distancetransform} to each segment individually. 
Illustrated in Fig.~\ref{fig_pipeline}, each pixel represents the least distance to its segment border, expressed as
$\mathbf{D} \in \mathbb{R}^{X \times Y \times 1}$:
\begin{equation}
    \mathbf{D} = \text{distance\_transform}\bigl(\mathbf{M}_{\text{safe}}\bigr) \; .
\end{equation}
Since the distances are measured on a metric scale (see Sec.~\ref{subsec_projection_registration}), we define an adjustable safety radius $r_{\text{safe}}$ that accounts for the physical UAV dimensions.
The segment with the largest distance value that exceeds this radius is chosen as the landing segment.
The coordinates of this maximum value define the center of the landing spot, denoted by $\mathbf{l} \in \mathbb{Z}^2$:

\begin{equation}
    \label{eq_l}
    \mathbf{l} = \argmax_{\mathbf{D}(x, y) \geq r_{\text{safe}}} \mathbf{D}(x, y) \; .
\end{equation}
All segments that satisfy these constraints are stored as a history of safe landing spots.
The ranking of semantic classes and the choice of the point of maximum distance within a safe segment are conservative design parameters that can be replaced by more adaptive criteria if required.
By utilizing a metric distance transform, we naturally account for UAV size constraints ($r_{\text{safe}}=$~\SI{3}{m} in experiments), enabling safe landings without prior map information, such as designed markers.

\subsection{Landing Procedure}\label{subsec_landing_logic}

\begin{figure}[!t]
    \centering
    \includegraphics[width=.75\linewidth]{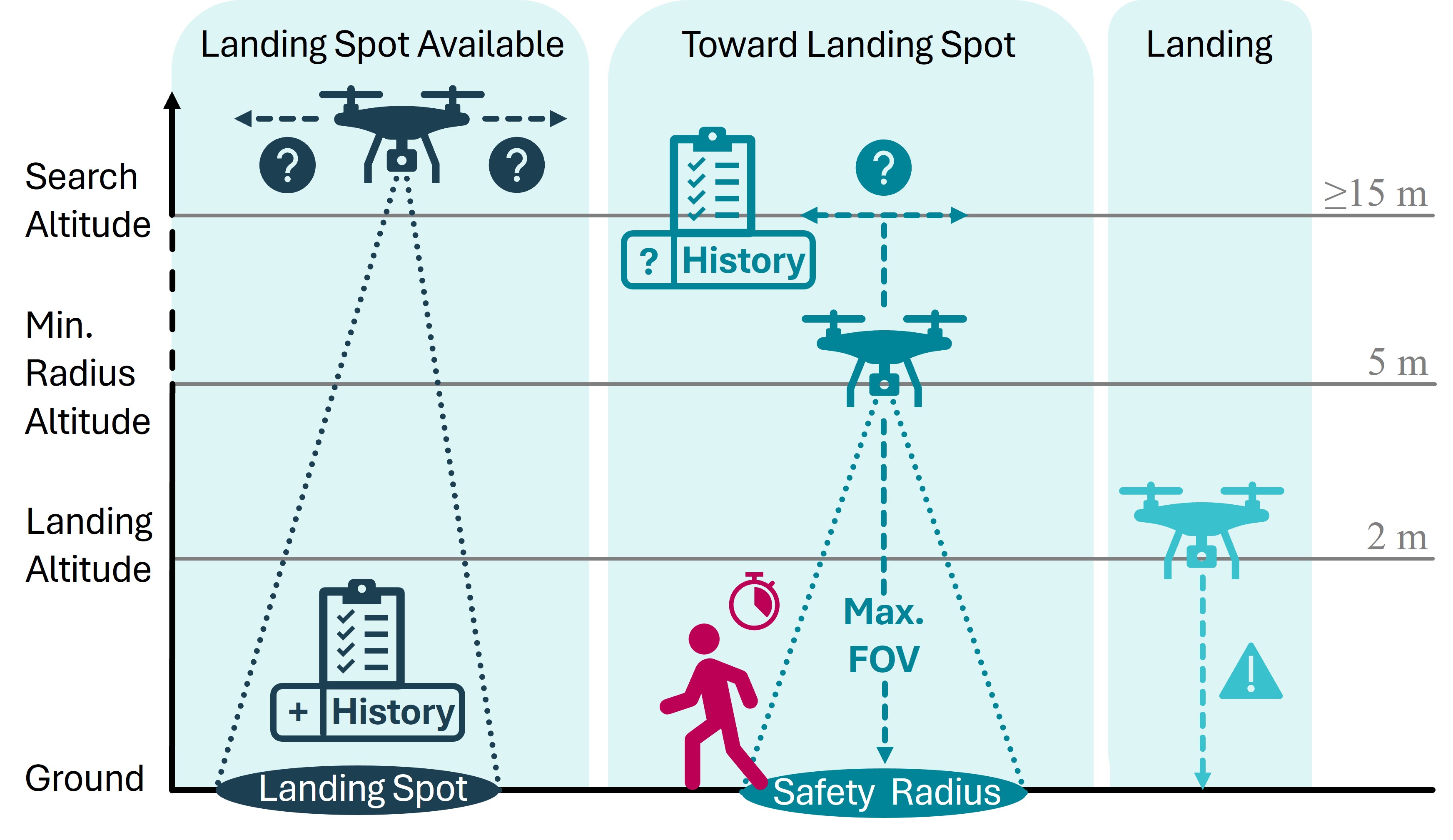}
    \caption{Landing procedure. A behavior tree (Sec.~\ref{subsec_landing_logic}) executes three distinct sequences that ensure safe, real-time landing in unknown and dynamic environments.}
    \label{fig_landing_logic}
\end{figure}

Once a safe landing spot $\mathbf{l}$ (Eqn.~\ref{eq_l}) has been identified, the landing procedure is governed by a behavior tree~\cite{behaviortree}, as illustrated in Fig.~\ref{fig_landing_logic}.
Behavior trees provide a modular and reactive decision-making framework, enabling continuous reassessment of the environment while maintaining deterministic execution.
This is particularly advantageous for safe autonomous landing, where environmental conditions may change throughout the descent and immediate reactions to dynamic obstacles are required.
The proposed behavior tree consists of three main execution sequences:\\

\noindent\textbf{Landing Spot Available Sequence.}
The first sequence continuously searches for valid landing locations while the UAV remains at or above an adjustable search altitude of \SI{15}{m} (Fig.~\ref{fig_landing_logic}).
During this phase, the complete perception pipeline, from semantic segmentation to landing spot detection (Sections~\ref{subsec_segmentation}--\ref{subsec_landing_spot_detection}), is executed repeatedly to maintain an up-to-date representation of the environment.
Whenever one or more valid landing spots are detected, their locations are stored in a history of validated landing sites.
This history enables the system to quickly revisit alternative candidates should the currently selected landing location become invalid later.
If no suitable landing site is available, the UAV continues along its current heading as a simple exploration strategy.
Since the search behavior is independent of the remaining landing pipeline, this can readily be replaced by more sophisticated mission-specific exploration or path-planning algorithms.\\

\noindent\textbf{Toward Landing Spot Sequence.}
Once a landing location has been selected, the UAV first performs level flight toward a waypoint directly above the landing spot before initiating its descent toward the predefined landing altitude (Fig.~\ref{fig_landing_logic}).
Importantly, selecting a landing site does not constitute a final commitment.
Instead, the system continuously re-evaluates the safety of the selected location throughout the descent using the perception pipeline.
If the landing site becomes invalid, e.g., due to newly observed terrain or the appearance of dynamic obstacles, the descent is immediately aborted, and the UAV climbs back to the search altitude.
From there, the system either identifies a new landing location or retrieves a previously validated candidate from the landing history.

This continuous reassessment is performed until the UAV reaches the minimum radius altitude of \SI{5}{m}.
At this point, the projected safety radius fully covers the camera's field of view (Fig.~\ref{fig_landing_logic}), eliminating the need to repeatedly evaluate the surrounding terrain.
Consequently, the system focuses exclusively on monitoring dynamic obstacles, as they represent the only remaining source of uncertainty.
Whenever a person or other non-cooperative obstacle enters the landing area, the UAV pauses its descent for up to \SI{5}{seconds}.
If the landing zone clears during this interval, the descent resumes immediately.
Otherwise, the landing attempt is aborted, and the UAV returns to the search altitude, where a new landing decision is initiated.\\

\noindent\textbf{Landing Sequence.}
The final sequence is entered once the UAV reaches the landing altitude of \SI{2}{m}.
At this stage, the UAV commits to the landing and no longer performs additional decision-making.
This design choice intentionally balances safety with flight stability.
At very low altitudes, aerodynamic ground effects become increasingly pronounced, while the semantic segmentation model exhibits reduced reliability below approximately \SI{2}{m} owing to the limited availability of low-altitude training imagery.
Moreover, by the time this altitude is reached, the selected landing site has already remained valid throughout the entire descent, substantially increasing confidence that the landing can be completed safely.
Consequently, introducing further decision points at this stage would provide limited benefit while unnecessarily increasing system complexity during the most critical phase of the maneuver.\\

Overall, the proposed behavior tree enables reactive, real-time landing decisions while remaining independent of prior maps or pre-defined landing infrastructure.
By continuously validating candidate landing sites and explicitly accounting for dynamic human activity until shortly before touchdown, the framework provides a practical and safety-oriented control strategy for autonomous landing in previously unseen environments.

\section{Experimental Results}
\label{sec_experimental_results}

\subsection{Simulation Experiments}\label{subsec_simulation_experiments}

\textbf{Setup:} We adopt a software-in-the-loop (SIL) setup using the OCTAS v1.0 simulation framework~\cite{octas}, integrated with ROS2 Humble~\cite{ros2} and PX4 v1.15.4~\cite{px4}, thereby closely mirroring the software stack employed during the real-world field experiments.
This enables us to evaluate the complete perception and landing pipeline under controlled yet realistic operating conditions while ensuring a seamless transition between simulation and hardware deployment.

Throughout the experiments, we define ``grass'' as the safe semantic class $\mathcal{C}_{\text{safe}}$ for landing.
This choice reflects the simulated environment but can readily be adapted to different semantic categories depending on mission-specific operational requirements.
Since the semantic segmentation model (Sec.~\ref{subsec_segmentation}) is trained exclusively on real-world imagery and is therefore not optimized for synthetic data, we use the simulator's semantic ground-truth annotations to isolate the evaluation of the proposed landing framework from the domain gap between simulated and real imagery.

To evaluate the robustness of the proposed approach under dynamic conditions, we perform 200 independent landing trials from an initial altitude of \SI{50}{m}.
During each trial, up to 10 non-cooperative dynamic obstacles, such as humans, are introduced to assess the system's ability to respond to changing environments by waiting or rerouting while maintaining safe landing execution.\\

\noindent\textbf{Results:} The first column of Fig.~\ref{fig_results} illustrates a representative simulation sequence.
The proposed Bayesian temporal-probabilistic filtering progressively refines the semantic scene representation, enabling the selection of a valid landing site within the predefined safe semantic classes despite continuously changing observations.
As the UAV descends, the selected landing location remains consistent while still adapting to newly observed information.

Whenever a non-cooperative obstacle entered the predefined safety radius, the behavior tree correctly suspended the landing procedure until the area became safe again or, if necessary, selected an alternative landing location.
Across all 200 trials, every landing was completed successfully without a single failed landing or incorrect rerouting decision.
These results demonstrate that the proposed framework reliably balances stable landing-site selection with rapid reactions to dynamic obstacles, providing a robust foundation before transferring the system to real-world deployments.

\subsection{Field Experiments}\label{subsec_field_experiments}

\begin{figure}[!t]
    \centering
    \includegraphics[width=.98\linewidth]{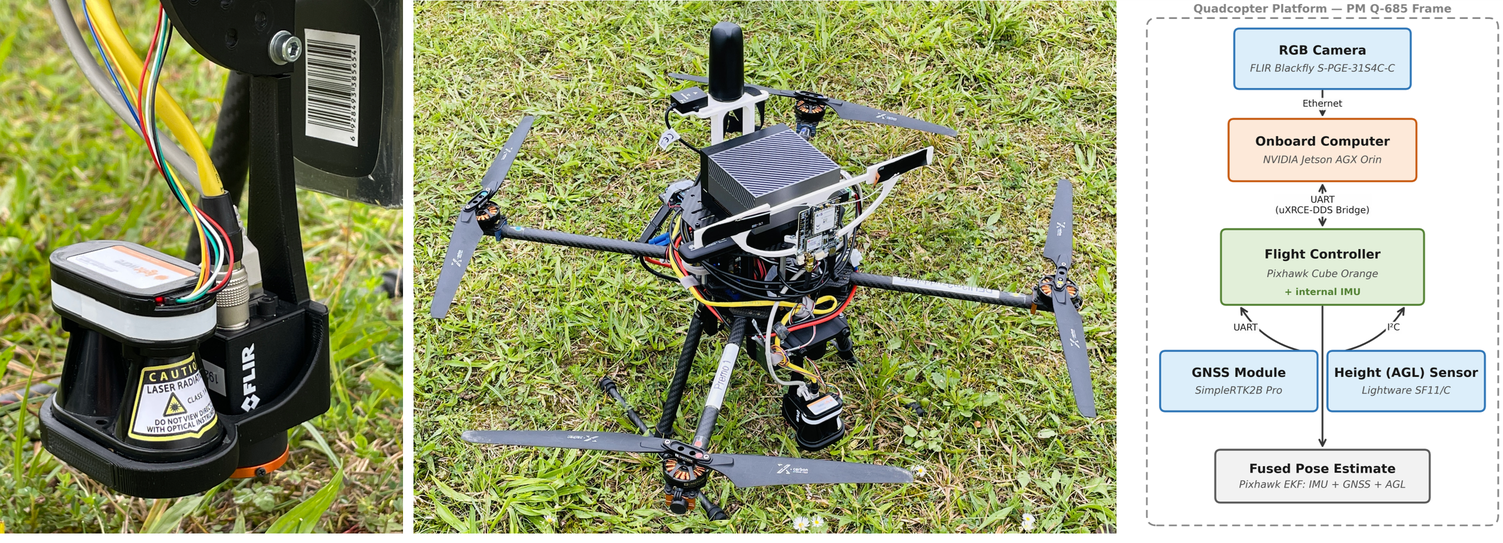}
    \caption{Hardware configuration. \textbf{Left}: Nadir-pointing RGB camera FLIR Blackfly S-PGE-31S4C-C \cite{flir} and the height AGL sensor Lightware SF11/C \cite{rangefinder}. \textbf{Center}: Quadcopter PM Q-685 \cite{premo} custom frame, equipped with NVIDIA Jetson AGX Orin \cite{orin}, GNSS module SimpleRTK2B Pro~\cite{gnss}, and flight controller Pixhawk Cube Orange with its IMUs~\cite{pixhwak}. \textbf{Right}: Block diagram of the \method~hardware setup.}
    \label{fig_hardware_config}
\end{figure}

\textbf{Setup:} To validate the proposed framework under real operating conditions, we implement the complete system in ROS2 Humble~\cite{ros2} using the hardware configuration shown in Fig.~\ref{fig_hardware_config}.
The onboard Jetson companion computer receives RGB images from the camera via Ethernet and communicates with the Pixhawk flight controller through a UART uXRCE-DDS Bridge~\cite{uxrce_bridge}.
The Pixhawk further interfaces with the GNSS receiver via UART and the above-ground-level (AGL) sensor through I2C.
Throughout all experiments, the UAV pose is obtained from the Pixhawk Extended-Kalman-Filter (EKF) estimator, which fuses measurements from the IMU, GNSS receiver, and AGL sensor.

The experimental campaign comprises a total of 60 autonomous landing flights conducted across three representative operational environments, including industrial, urban, and rural scenes, with 20 flights performed in each environment.
To evaluate robustness across different mission profiles, flights are conducted from altitudes of \SI{25}{m}, \SI{50}{m}, and \SI{100}{m}.
We define ``grass'' and ``gravel'' as safe semantic categories $\mathcal{C}_{\text{safe}}$, although these semantic definitions can be readily adapted to mission-specific requirements.
To evaluate the safety mechanisms of the proposed framework, up to three participants actively enter the designated landing zone during the landing procedure, forcing the system to react to dynamic human presence.\\

\begin{figure*}[t!]
	\centering
	\includegraphics[width=.98\textwidth]{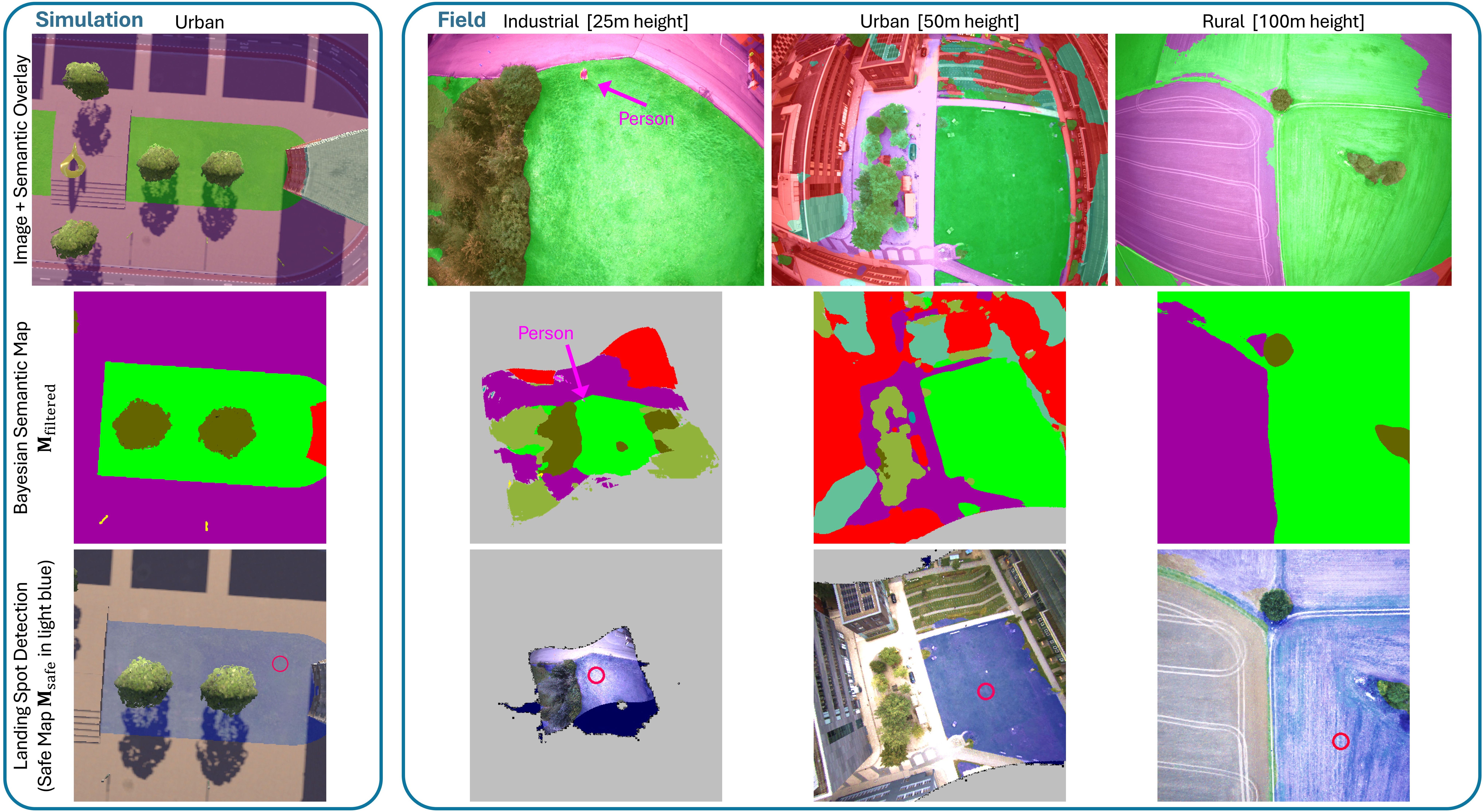}
	\caption{Experimental results (Sec.~\ref{sec_experimental_results}) in 200 simulations and 60 industrial, urban, and rural field trials ($25$$-$\SI{100}{m} AGL) show that \method~reliably identifies metric-scale landing sites and responds to non-cooperative humans in real time, achieving \SI{95}{\%} end-to-end landing success, zero false negatives for human detection, and sub-second response latency. Note that the map has a fixed, metric-scale size, resulting in a smaller map at lower heights, as the camera field of view covers less area.}
    \label{fig_results}    
\end{figure*}

\noindent\textbf{Results:} Representative examples of the field experiments are shown in Fig.~\ref{fig_results}.
Out of the 60 autonomous landing attempts, 57 were completed successfully, corresponding to a landing success rate of \SI{95}{\%}.
The remaining three flights were safely aborted due to transient communication failures between the Pixhawk flight controller and the Jetson companion computer, which are external to the proposed perception and decision-making framework.

The experiments further demonstrate reliable operation in the presence of dynamic obstacles.
Whenever individuals entered the landing zone, the behavior tree either paused the descent until the area became safe again or rerouted the UAV to an alternative landing location when the originally selected site remained occupied.
During two flights, the semantic segmentation model temporarily produced false-positive human detections within the landing zone, resulting in conservative delays in the landing procedure.
These false detections were subsequently corrected through Bayesian temporal filtering and semantic decay as additional observations became available, thereby allowing both landings to be completed successfully.
Importantly, no false negatives occurred throughout the entire experimental campaign, meaning that every person entering the landing zone was successfully detected.
This conservative behavior highlights the safety-oriented design of the proposed framework, in which temporary delays are preferred over potentially unsafe landing decisions.

Across all evaluated environments and flight altitudes, the semantic segmentation model consistently identified suitable landing regions under varying scene layouts and illumination conditions.
Combined with Bayesian filtering and semantic decay, the resulting semantic world representation remained both stable and responsive, enabling the behavior tree to make reliable landing decisions while continuously accounting for dynamic human activity.
Running on the onboard Jetson, the complete perception and decision-making pipeline achieves end-to-end processing rates of \SI{1.3}{Hz} (\SI{30}{W}), \SI{2.0}{Hz} (\SI{50}{W}), and \SI{3.2}{Hz} (MAXN mode), demonstrating that the proposed system satisfies the computational requirements for real-time onboard deployment.

\subsection{Baseline Comparison}\label{subsec_comparison}

\begin{table}[t!]
    \centering
    \caption{Baseline comparison of autonomous landing methods against requirements identified in recent survey literature~\cite{sssurvey,farajijalal2025safety} (see Sec.~\ref{subsec_comparison}). \textbf{--} marks ambiguous references.}
    \label{tab_related_word}
    \begin{adjustbox}{max width=.5\linewidth}
    \begin{tabular}{l|c|c|c|c|c}
        \toprule
        Reference & \begin{tabular}{c}No \\ Prior\end{tabular} & Metric & \begin{tabular}{c}No \\ Lid./Rad.\end{tabular} & People & Field \\ 
        \midrule
        \cite{hub_1,hub_1_2,neves2024} & \textcolor{red}\xmark & \cmark & \textcolor{red}\xmark & \textcolor{red}\xmark & \textcolor{red}\xmark \\
        \cellcolor{rowcolor}\cite{hub_2} & \cellcolor{rowcolor}\textcolor{red}\xmark & \cellcolor{rowcolor}\textcolor{red}\xmark & \cellcolor{rowcolor}\cmark & \cellcolor{rowcolor}\textcolor{red}\xmark & \cellcolor{rowcolor}\textcolor{red}\xmark \\
        \cite{multimodal} & \textcolor{red}\xmark & \cmark & \textbf{--} & \textcolor{red}\xmark & \textcolor{red}\xmark \\
        \cellcolor{rowcolor}\cite{landing_target,russian_paper} & \cellcolor{rowcolor}\textcolor{red}\xmark & \cellcolor{rowcolor}\cmark & \cellcolor{rowcolor}\cmark & \cellcolor{rowcolor}\textcolor{red}\xmark & \cellcolor{rowcolor}\textcolor{red}\xmark \\
        \cite{safeeye_1, safeeye_2,monodepthEML} & \cmark & \textbf{--} & \cmark & \textcolor{red}\xmark & \textcolor{red}\xmark \\
        \cellcolor{rowcolor}\cite{satellite} & \cellcolor{rowcolor}\cmark & \cellcolor{rowcolor}\textcolor{red}\xmark & \cellcolor{rowcolor}\cmark & \cellcolor{rowcolor}\textcolor{red}\xmark & \cellcolor{rowcolor}\textcolor{red}\xmark \\
        \cite{safeeye_3,gabor,SORA} & \cmark & \textcolor{red}\xmark & \cmark & \textcolor{red}\xmark & \textcolor{red}\xmark \\
        \cellcolor{rowcolor}\cite{lidar_1,lidar_2} & \cellcolor{rowcolor}\cmark & \cellcolor{rowcolor}\cmark & \cellcolor{rowcolor}\textcolor{red}\xmark & \cellcolor{rowcolor}\textcolor{red}\xmark & \cellcolor{rowcolor}\textcolor{red}\xmark \\
        \cite{pakistani} & \cmark & \cmark & \textbf{--} & \textcolor{red}\xmark & \textcolor{red}\xmark \\
        \cellcolor{rowcolor}\cite{canny_land, free_1,freelsd,luczak2025autonomous,11247278} & \cellcolor{rowcolor}\cmark & \cellcolor{rowcolor}\cmark & \cellcolor{rowcolor}\cmark & \cellcolor{rowcolor}\textcolor{red}\xmark & \cellcolor{rowcolor}\textcolor{red}\xmark \\
        \cite{PIECZYNSKI2024107864,elpodense} & \textcolor{red}\xmark & \cmark & \cmark & \cmark & \textcolor{red}\xmark \\
        \cellcolor{rowcolor}\cite{10971405} & \cellcolor{rowcolor}\cmark & \cellcolor{rowcolor}\cmark & \cellcolor{rowcolor}\textcolor{red}\xmark & \cellcolor{rowcolor}\textcolor{red}\xmark & \cellcolor{rowcolor}\cmark \\
        \cite{circles_1} & \cmark & \textcolor{red}\xmark & \cmark & \cmark & \textcolor{red}\xmark \\
        \cellcolor{rowcolor}\cite{circles_2, circles_3,ESLS,De_La_Torre_Vanegas_2026} & \cellcolor{rowcolor}\cmark & \cellcolor{rowcolor}\cmark & \cellcolor{rowcolor}\cmark & \cellcolor{rowcolor}\cmark & \cellcolor{rowcolor}\textcolor{red}\xmark \\
        \cellcolor{lightorange}Safe2Ditch \cite{safe2ditch} & \cellcolor{lightorange}\textcolor{red}\xmark & \cellcolor{lightorange}\cmark & \cellcolor{lightorange}\cmark & \cellcolor{lightorange}\cmark & \cellcolor{lightorange}\cmark \\
        \cellcolor{lightgreen}\method~(ours) & \cellcolor{lightgreen}\cmark & \cellcolor{lightgreen}\cmark & \cellcolor{lightgreen}\cmark & \cellcolor{lightgreen}\cmark & \cellcolor{lightgreen}\cmark \\
        \bottomrule
    \end{tabular}
    \end{adjustbox}
\end{table}

Existing work on safe autonomous landing (Sec.~\ref{sec_related_work}) together with recent comprehensive surveys~\cite{sssurvey,farajijalal2025safety} identifies several fundamental requirements for practical vision-based landing systems.
Such systems should (1) operate without prior information about candidate landing zones to enable deployment in unknown and dynamic environments, (2) estimate landing areas in metric units to account for different UAV dimensions, (3) rely on lightweight sensing modalities such as RGB cameras rather than heavier and more power-intensive sensors including LiDAR or radar, (4) explicitly prioritize human safety throughout the landing procedure, and (5) demonstrate complete end-to-end validation through real-world flight experiments.
Table~\ref{tab_related_word} summarizes these requirements and shows that, to the best of our knowledge, our method is the first to satisfy all of them simultaneously.\\

Among existing approaches, Safe2Ditch~\cite{safe2ditch} is the only work that presents a complete autonomous landing system together with comprehensive field validation.
We therefore consider it the strongest available baseline, despite relying on a database of pre-selected landing sites as prior information.
This assumption fundamentally differs from our problem formulation, which deliberately avoids prior environmental knowledge to enable deployment in previously unseen environments.\\

To evaluate responsiveness in dynamic environments, we compare the response latency, defined as the time between a non-cooperative obstacle entering the currently selected landing zone and the triggering of a safety intervention.
This metric directly reflects the ability of a landing system to react to unforeseen hazards during the critical final landing phase, where delayed decisions reduce the available altitude and maneuvering time.
Since Safe2Ditch reports response latencies only graphically, we digitize the published plots to recover the reported measurements.
To ensure a fair comparison, we apply an identical aggregation procedure to both methods by averaging simulation results across scenarios containing one to ten non-cooperative obstacles and averaging field measurements across all reported flight altitudes.

\begin{figure}[t!]
    \centering
    \includegraphics[width=.85\linewidth]{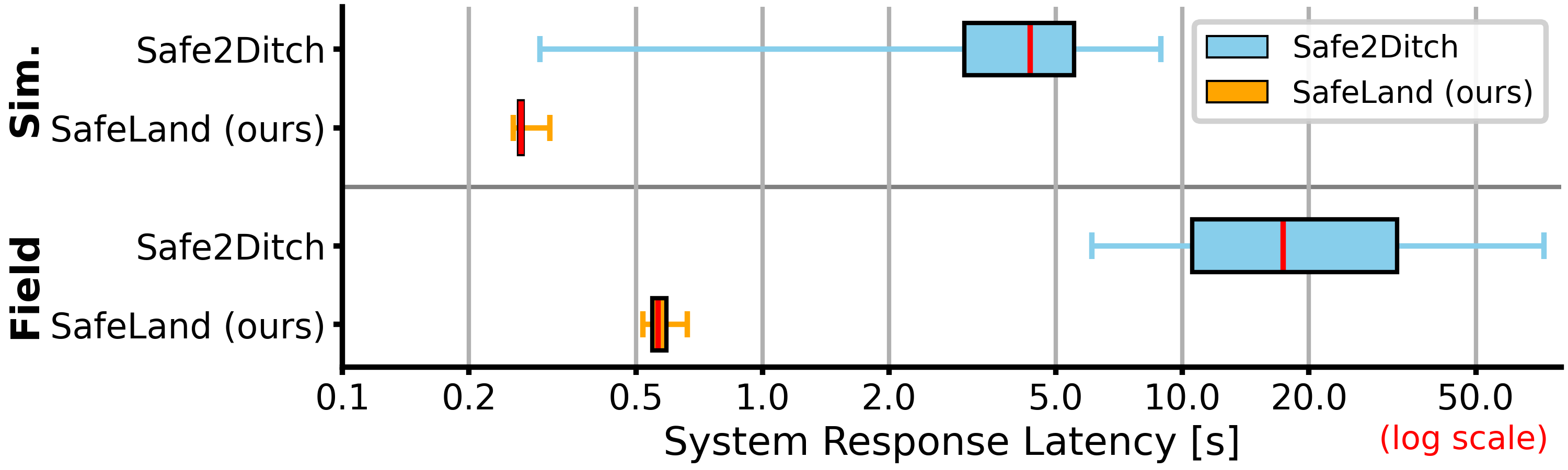}
    \caption{System response latencies (see Sec.~\ref{subsec_comparison}) on dynamic obstacles for Safe2Ditch~\cite{safe2ditch} and \method~(ours). We substantially outperform Safe2Ditch by providing sub-second response latency as a critical capability for unknown and dynamic environments.}
    \label{fig_response_latency}
\end{figure}

Figure~\ref{fig_response_latency} shows that our method consistently achieves substantially lower response latency, providing sub-second reactions to dynamic obstacles.
This improvement follows directly from the architectural differences between the two systems.
Safe2Ditch evaluates dynamic obstacles with respect to a database of pre-selected landing sites, requiring tracked objects to be geolocated before rerouting decisions can be made.
Safe2Ditch identifies this target geolocation stage as the dominant contributor to the overall response latency due to localization uncertainty.
In contrast, our method performs safety assessment directly on the current semantic scene representation and therefore does not require prior landing-site databases or an intermediate target geolocation stage.
Besides enabling operation in unknown environments, this architectural simplification substantially reduces the system response time.\\

Finally, we compare failure rates.
In simulation, both methods achieve effectively zero failures.
During field experiments, Safe2Ditch reports one failure in sixteen flights caused by obstacle misclassification, whereas our system experienced three failures in sixty flights, corresponding to a success rate of 95\%.
Importantly, all observed failures were traced to transient communication issues between the Pixhawk flight controller and the Jetson companion computer and are therefore independent of the proposed perception and landing framework.

\section{Conclusion}
\label{sec_conclusion}

In this work, we presented a lightweight vision-based framework for safe autonomous landing in unknown and dynamic environments.
By integrating semantic segmentation, Bayesian temporal-probabilistic filtering, and behavior tree-based decision making, the proposed system enables autonomous landing site selection and execution without relying on prior maps, predefined landing hubs, or heavy perception sensors, while explicitly considering human safety.
Extensive simulation and real-world evaluations demonstrate the effectiveness of the proposed approach, achieving a \SI{95}{\%} landing success rate, sub-second responses to dynamic obstacles, and zero false negatives in human detection. To facilitate further research in vision-based aerial robotics, we release our segmentation model as a plug-and-play ROS2 package.\\

Future work will focus on relaxing current assumptions by incorporating terrain-aware uncertainty modeling, improving robustness under challenging environmental conditions, and leveraging recent advances in metric depth estimation and large-scale aerial segmentation, such as OccuFly~\cite{gross2026occufly} and SegFly~\cite{gross2026segfly}, to replace dedicated height sensing and extend semantic understanding beyond the current system capabilities.

\bibliographystyle{splncs04}
\bibliography{main}

\end{document}